\documentclass[conference]{IEEEtran}
\IEEEoverridecommandlockouts

\ifCLASSINFOpdf
\else
\fi

\usepackage{fancyhdr}
\usepackage[dvips]{graphicx}
\usepackage{indent}
\usepackage{amsmath,amssymb}
\usepackage{paralist}
\usepackage{eclbkbox}
\usepackage{subfigure}
\usepackage{url}

\makeatletter
\def\tbcaption{\def\@captype{table}\caption}
\def\figcaption{\def\@captype{figure}\caption}

\newcommand{\bvec}[1]{\mbox{\boldmath $#1$}}

\begin{document}
\title{An Adaptive Learning Method of\\ Restricted Boltzmann Machine by\\ Neuron Generation and Annihilation Algorithm
\thanks{\copyright 2016 IEEE. Personal use of this material is permitted. Permission from IEEE must be obtained for all other uses, in any current or future media, including reprinting/republishing this material for advertising or promotional purposes, creating new collective works, for resale or redistribution to servers or lists, or reuse of any copyrighted component of this work in other works.}}

\author{
\IEEEauthorblockN{Shin Kamada}
\IEEEauthorblockA{
Graduate School of Information Sciences, \\
Hiroshima City University\\
3-4-1, Ozuka-Higashi, Asa-Minami-ku,\\
Hiroshima, 731-3194, Japan\\
Email: da65002@e.hiroshima-cu.ac.jp}
\and
\IEEEauthorblockN{Takumi Ichimura}
\IEEEauthorblockA{Faculty of Management and Information Systems,\\
Prefectural University of Hiroshima\\
1-1-71, Ujina-Higashi, Minami-ku,\\
Hiroshima, 734-8559, Japan\\
Email: ichimura@pu-hiroshima.ac.jp}
}

\maketitle
\fancypagestyle{plain}{
\fancyhf{}	
\fancyfoot[L]{}
\fancyfoot[C]{}
\fancyfoot[R]{}
\renewcommand{\headrulewidth}{0pt}
\renewcommand{\footrulewidth}{0pt}
}

\pagestyle{fancy}{
\fancyhf{}
\fancyfoot[R]{}}
\renewcommand{\headrulewidth}{0pt}
\renewcommand{\footrulewidth}{0pt}

\begin{abstract}
  Restricted Boltzmann Machine (RBM) is a generative stochastic energy-based model of artificial neural network for unsupervised learning. Recently, RBM is well known to be a pre-training method of Deep Learning. In addition to visible and hidden neurons, the structure of RBM has a number of parameters such as the weights between neurons and the coefficients for them. Therefore, we may meet some difficulties to determine an optimal network structure to analyze big data. In order to evade the problem, we investigated the variance of parameters to find an optimal structure during learning. For the reason, we should check the variance of parameters to cause the fluctuation for energy function in RBM model. In this paper, we propose the adaptive learning method of RBM that can discover an optimal number of hidden neurons according to the training situation by applying the neuron generation and annihilation algorithm. In this method, a new hidden neuron is generated if the energy function is not still converged and the variance of the parameters is large. Moreover, the inactivated hidden neuron will be annihilated if the neuron does not affect the learning situation. The experimental results for some benchmark data sets were discussed in this paper. 
\end{abstract}

\IEEEpubidadjcol

\section{Introduction}
\label{sec:Introduction}
The current information technology can collect various kinds of data sets, because the recent tremendous technical advances in processing power, storage capacity, and network connected to cloud computing. Such data sample includes not only numerical values but also text such as comments, numerical evaluation such as ranking, and binary data such as pictures. Such data set is called big data. The technical methods to discover knowledge from big data are known to be a field of data mining and also developed in the research field of Deep Learning \cite{Quoc12}.

Deep Learning attracts a lot of attention in methodology research of artificial intelligence such as machine learning \cite{Bengio09}. Especially, the industrial world is deeply impressed by the outcome to increase the capability of image processing. The learning architecture has an advantage of not only multi-layered network structure but also pre-training. The latter characteristic means that
the architecture of Deep Learning accumulates prior knowledge of the features for input patterns. Restricted Boltzmann Machine (RBM) \cite{Hinton12} is one of popular method of Deep Learning for unsupervised learning. RBM has the capability of representing an probability distribution of input data set, and it can represent an energy-based statistical model. Moreover, the Contrastive Divergence (CD) learning procedure which is a faster algorithm of Gibbs sampling based on Markov chain Monte Carlo methods can be often used as one of the learning methods of RBM \cite{Hinton02,Tileman08}. 

The problem related to RBM is how to determine the definition of an optimal initial network structure such as the number of hidden neurons according to the features of input pattern because the traditional RBM model cannot change its network structure during learning phase. In this paper, we propose the adaptive learning method of RBM that can discover an optimal number of hidden neurons according to the training situation by applying the neuron generation and annihilation algorithm. In multi-layered neural networks, the adaptive learning method by neuron generation and annihilation algorithm during learning phase was proposed \cite{Ichimura97,Ichimura04}. The method monitors the variance of the weight vectors called the Walking Distance ($WD$) in the learning phase. A new neuron will be generated and inserted into the related position if the weight vector tends to fluctuate greatly even after a certain period of the training process. Moreover, the inactivated hidden neuron will be annihilated if the neuron does not affect the learning situation.

However, RBM with CD method works in an output to use binary neurons. We consider to convergence under the Lipschitz continuous condition \cite{Carlson15}. According to \cite{Carlson15}, the energy function of RBM can be transformed to the equations under the continuous conditions with 3 kinds of parameters for visible and hidden neurons. We investigated the variance of 3 kinds parameters where energy function of RBM converges \cite{Kamada15a}. Then we selected 2 parameters which influence the convergence situation of RBM except the parameter related to input features. In this paper, we show that our proposed model has the good classification capability for the small data set (about 1000 records \cite{Kamada15b}). Moreover, we applied our proposed adaptive learning method of RBM to big data set such as CIFAR-10 \cite{cifar10}. From experimental results, our proposed model will be the good performance in comparison to previous RBM model \cite{Dieleman12}.

The remainder of this paper is organized as follows. Section \ref{sec:RBM} describes the basic concept of RBM and the condition of convergence under the Lipschitz continuous is derived. In Section \ref{subsec:WD}, neuron generation and annihilation algorithm in multi-layered neural networks is explained and we apply this method into RBM in Section \ref{subsec:AdaptiveRBM}. Section \ref{sec:EXE} describes some experimental results. We give some discussions to conclude this paper in Section \ref{sec:Conclusion}.

\section{Restricted Boltzmann Machine}
\label{sec:RBM}

\subsection{Overview}
\begin{figure}[tbp]
\begin{center}
\includegraphics[scale=0.6]{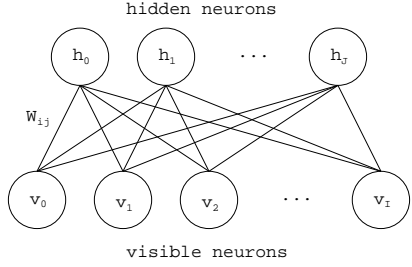}
\vspace{-3mm}
\caption{The structure of RBM}
\vspace{-3mm}
\label{fig:rbm}
\end{center}
\end{figure}

This section explains the basic concept of RBM \cite{Hinton12}. As shown in Fig.\ref{fig:rbm}, RBM has the network structure with 2 kinds of layers where one is a visible layer for input data and the other is a hidden layer for representing the features of given data space. Each layer consists of some binary neurons. The traditional Boltzmann Machine has the connections between neurons in the same layer \cite{Ackley85}. However RBM has no connection in the same layer. Therefore the calculation is easier than the traditional one from the viewpoint of the no interaction between neurons. The RBM learning employs to train the weights and some parameters for visible and hidden neurons till the energy function becomes to a certain small value. The trained RBM can represent a probability for the distribution of input data.

Let $v_i (0 \leq i \leq I)$ and $h_j (0 \leq j \leq J)$ be binary variable of a visible neuron and a hidden neuron, respectively. $I$ and $J$ are the number of visible and hidden neurons, respectively. The energy function $E(\bvec{v}, \bvec{h})$ for visible vector $\bvec{v} \in \{ 0, 1 \}^{I}$ and hidden vector $\bvec{h} \in \{ 0, 1 \}^{J}$ is given by Eq.(\ref{eq:energy}). $p(\bvec{v}, \bvec{h})$ is the joint probability distribution of $\bvec{v}$ and $\bvec{h}$ as shown in Eq.(\ref{eq:prob}).

\vspace{-2mm}
\begin{equation}
E(\bvec{v}, \bvec{h}) = - \sum_{i} b_i v_i - \sum_j c_j h_j - \sum_{i} \sum_{j} v_i W_{ij} h_j ,
\label{eq:energy}
\end{equation}

\vspace{-2mm}
\begin{equation}
p(\bvec{v}, \bvec{h})=\frac{1}{Z} \exp(-E(\bvec{v}, \bvec{h})) ,
\label{eq:prob}
\end{equation}

\vspace{-2mm}
\begin{equation}
Z = \sum_{\bvec{v}} \sum_{\bvec{h}} \exp(-E(\bvec{v}, \bvec{h})) ,
\label{eq:PartitionFunction}
\end{equation}
where $b_i$ and $c_j$ are the parameters for $v_i$ and $h_j$, respectively. $W_{ij}$ is the weight between $v_i$ and $h_j$. $Z$ is the partition function which is given by summing over all possible pairs of visible and hidden vectors. 

The parameters of RBM are updated by maximum likelihood estimation for $p(\bvec{v})=\sum_{\bvec{h}} p(\bvec{v}, \bvec{h})$ which is the probability of $\bvec{v}$. However, the computational elements increase exponentially because the optimal configuration for all possible pairs is required to obtain the maximum likelihood estimation. Therefore, the Contrastive Divergence (CD) learning procedure has been proposed as RBM training. CD method can be a faster algorithm of Gibbs sampling based on Markov chain Monte Carlo methods \cite{Hinton02}. Then CD method is known to make a good performance even in a few sampling steps \cite{Tileman08}.

\subsection{Convergence under the Lipschitz continuous condition \cite{Carlson15}}
\label{subsec:RBM_bound}
CD method works in discrete space. Therefore, we consider the convergence situation of RBM under the Lipschitz continuous condition. Generally, the solution will be found by using machine learning if and only if the convexity and continuous conditions for an objective function are satisfied. However RBM learning with CD sampling method meets the situation that may cause the slight error and it may not satisfy the continuous condition because of the use of binary neuron. Even if the network has a small error in the initial step, but the total energy after a certain period of iterations will be fluctuated seriously.

Carlson et al. discussed the upper bounds on the log partition function for each parameter of RBM by the convexity and Lipschitz continuous \cite{Carlson15}. The paper derived the following equations to measure the likelihood for 3 kinds of parameters $\bvec{\theta}=\{\bvec{b}, \bvec{c}, \bvec{W} \}$.

\vspace{-2mm}
\begin{equation}
E(\bvec{v},\bvec{h};\bvec{\theta})=-\sum_{i} v_i b_i - \sum_j h_j c_j - \sum_{i} \sum_{j} v_i W_{ij} h_j,
\label{eq:energy2}
\end{equation}
\vspace{-2mm}
\begin{equation}
\arg \min_{\bvec{\theta}} F(\bvec{\theta}) = \frac{-1}{N}\sum_{n=1}^{N}{\log} p_{\bvec{\theta}}(\bvec{v}_{n}) = f(\bvec{\theta})-g(\bvec{\theta}),
\label{eq:log_likelihood}
\end{equation}
\vspace{-2mm}
\begin{equation}
f(\bvec{\theta})={\rm log}\sum_{\bvec{v}} \sum_{\bvec{h}}{\exp}(-E(\bvec{v},\bvec{h};\bvec{\theta})),
\label{eq:f_theta}
\end{equation}
\vspace{-2mm}
\begin{equation}
g(\bvec{\theta})=\frac{1}{N}\sum_{n=1}^{N}{\rm log}\sum_{\bvec{h}}{\rm exp}(-E(\bvec{v},\bvec{h};\bvec{\theta})) ,
\label{eq:g_theta}
\end{equation}
where $\bvec{v}_{n}=\{ \bvec{v}_{1},\bvec{v}_{2},\cdots,\bvec{v}_{N}\}$ is a given input data, $N$ is the number of samples of input data. Eq.(\ref{eq:f_theta}) and Eq.(\ref{eq:g_theta}) are log likelihood functions for ideal model and real model for input data, respectively. $f(\bvec{\theta})$ is the log partition function and it is estimated by the sampling method such as CD method. $f(\bvec{\theta})$ can be transformed to Eq.(\ref{eq:upper_bound_f_theta_b}), Eq.(\ref{eq:upper_bound_f_theta_c}) and Eq.(\ref{eq:upper_bound_f_theta_W}) for each parameter under the assumption of Lipschitz continuous (see \cite{Carlson15} for details). As a result, these equations can be derived from the Taylor's expansion under the partial derivative of $f(\bvec{\theta})$ in each parameter.
\vspace{-3mm}
\begin{equation}
f(\{ \bvec{b}, \bvec{c}^{k}, \bvec{W}^{k} \}) \leq f(\bvec{\theta}^{k}) + \langle \nabla_{\bvec{b}} f(\bvec{\theta}^{k}), \bvec{b} - \bvec{b}^{k} \rangle + \frac{I}{2} \| \bvec{b} - \bvec{b}^{k} \|^{2}_{\infty},
\label{eq:upper_bound_f_theta_b}
\end{equation}
\vspace{-3mm}
\begin{equation}
f(\{ \bvec{b}^{k}, \bvec{c}, \bvec{W}^{k} \}) \leq f(\bvec{\theta}^{k}) + \langle \nabla_{\bvec{c}} f(\bvec{\theta}^{k}), \bvec{c} - \bvec{c}^{k} \rangle + \frac{J}{2} \| \bvec{c} - \bvec{c}^{k} \|^{2}_{\infty},
\label{eq:upper_bound_f_theta_c}
\end{equation}
\vspace{-3mm}
\begin{eqnarray}
f(\{ \bvec{b}^{k}, \bvec{c}^{k}, \bvec{W} \}) &\leq& f(\bvec{\theta}^{k}) + tr( (\bvec{W} - \bvec{W}^{k})^{T} \nabla_{\bvec{W}} f(\bvec{\theta}^{k})) \nonumber \\
 &&  + \frac{2IJ}{2} \| \bvec{W} - \bvec{W}^{k} \|^{2}_{S^{\infty}},
\label{eq:upper_bound_f_theta_W}
\end{eqnarray}
where $I$ and $J$ are the number of visible neurons and hidden neurons, respectively. $S^{\infty}$ is Shatten-norm. $\langle \bvec{a}, \bvec{b} \rangle $ means an inner product between 2 vectors of $\bvec{a}$ and $\bvec{b}$. The upper bound equations for each parameter are based on the definition of Lipschitz continuous condition, that is the third term of right side of each equation means the range of learning convergence. Therefore, the RBM learning by CD method will be converged if the variance for each parameter falls into a certain range during training.

\begin{figure}[tbp]
\begin{center}
  \includegraphics[scale=0.65]{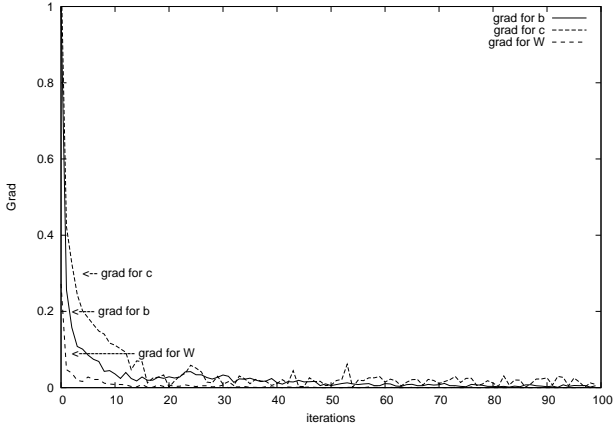}
\vspace{-3mm}
\caption{Gradient for b, c and W}
\label{fig:result-rbm-mnist-b-c-W-10}
\vspace{-3mm}
\end{center}
\end{figure}

We investigated the change of gradients for 3 kinds of parameters $\bvec{\theta}=\{\bvec{b}, \bvec{c}, \bvec{W} \}$ for the data set MNIST \cite{mnist}. Fig.\ref{fig:result-rbm-mnist-b-c-W-10} shows the gradients for each parameter in the simulation result. As shown in Fig.\ref{fig:result-rbm-mnist-b-c-W-10}, the gradients for each parameter became gradually small. The gradient for parameter $\bvec{c}$ was fluctuated large in 3 kinds of parameter. Moreover, the gradient for $\bvec{W}$ was changed according to the relationship between $\bvec{b}$ and $\bvec{c}$. On the other hand, the gradient for $\bvec{b}$ was also changed in the simulation result. However we found the gradient for parameter $\bvec{b}$ may be fluctuated according to the features of input patterns because parameter $\bvec{b}$ is the bias for input space \cite{Kamada15a}. Therefore, we selected 2 parameters which have an influence on the convergence situation of RBM except the parameter related to input data.

\section{Adaptive Learning Method of Restricted Boltzmann Machine}
\label{sec:AdaptiveRBM}

\subsection{Walking Distance in Multi-Layered Neural Network \cite{Ichimura97,Ichimura04}}
\label{subsec:WD}
\begin{figure}[tp]
\begin{center}
\includegraphics[scale=0.9]{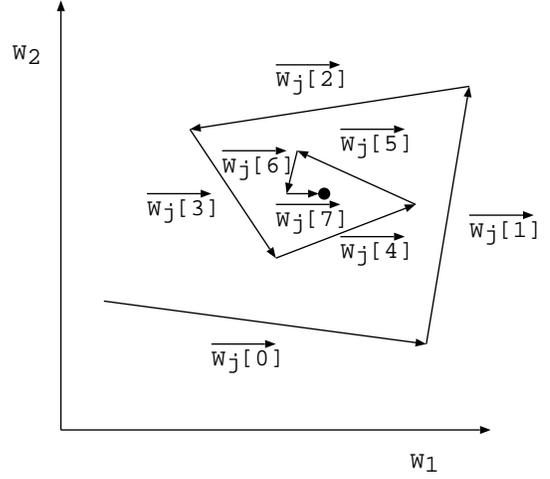}
\vspace{-3mm}
\caption{An image of convergence of a weight vector}
\label{fig:WD}
\vspace{-3mm}
\end{center}
\end{figure}

The problem related to the search of an optimal number of hidden neurons has also been considered in multi-layered neural networks. The neuron generation and annihilation algorithm during learning phase was proposed \cite{Ichimura97,Ichimura04}. Typically, a hidden neuron tries to learn input features by mapping original input data into feature vector. If a neural network does not have enough neurons to be satisfied to infer, then an input weight vector will tend to fluctuate greatly even after a certain period of the training process, because some hidden neurons may not represent an ambiguous patterns due to lack of the number of hidden neurons. In such a case, we can solve this problem by dividing a neuron which tries to represent the ambiguous patterns into 2 neurons by inheriting the attributes of the parent hidden neuron. The process is called the neuron generation. After an optimal number of neurons are generated, the network will be more stable within a small error. Therefore, we monitor the variance of the weight vector called the Walking Distance ($WD$) in hidden neurons while training as shown in Fig.~\ref{fig:WD}. $WD$ of a hidden neuron $j$ after $m$ training iterations is approximated by Eq.(\ref{eq:WD}).

\vspace{-3mm}
\begin{equation}
WD_j[m] = \gamma_{w} WD_{j} [m-1] + (1 - \gamma_{w}) Met(\vec{W}_j[m], \vec{W}_j[m-1]) ,   
\label{eq:WD}
\end{equation}

\noindent
where $\vec{W}_j[m]$ is a weight vector of a hidden neuron $j$ after $m$ training iterations, $Met$ is a metric function such as Euclidean Distance. $\gamma_{w}$ is a constant value in $[0,1]$. A new neuron is generated and is inserted into the neighborhood of the neuron $j$ when $WD_j$ is larger than the certain threshold as following equation \cite{Ichimura97,Ichimura04}.
\vspace{-3mm}
\begin{equation}
\Delta \varepsilon_j = \frac{\partial \varepsilon}{\partial WD_j} \cdot WD_j,
\label{eq:delta_epsilon}
\end{equation}
where $\varepsilon$ is the sum of squared error of a network.

On the other hand, if a neural network has enough neurons to infer and even if an input weight vector of each neuron converges smaller than a certain value, we shall be able to find unnecessary neurons from the network. In such a case, we proposed the neuron annihilation algorithm which can annihilate the redundant neuron if the variance of output signal for a neuron is smaller than a certain threshold as following equations.

\vspace{-3mm}
\begin{equation}
VA_j[m]  = \gamma_{v} VA_{j} [m-1] + (1 - \gamma_{v}) (O_j - Act_j [m])^{2} ,
\label{eq:Annihilation}
\end{equation}

\vspace{-3mm}
\begin{equation}
Act_j[m] = \gamma_{a} Act_{j} [m-1] + (1 - \gamma_{a}) O_j ,
\label{eq:Annihilation}
\end{equation}
where $O_j$ is the variance of the output signal for a neuron $j$. $\gamma_{v}$ and $\gamma_{a}$ are constant values in $[0,1]$.

\subsection{Neuron Generation and Annihilation Algorithm in RBM}
\label{subsec:AdaptiveRBM}
We propose the adaptive learning method of RBM that can discover an optimal number of hidden neurons by applying the neuron generation and annihilation algorithm. However, the structure of RBM has 3 kinds of parameters for visible and hidden neurons in addition to the weights between them. As mentioned in the section \ref{subsec:RBM_bound}, the RBM learning will be converged if the third term of right side in Eq.(\ref{eq:upper_bound_f_theta_b}) - (\ref{eq:upper_bound_f_theta_W}) become small. We consider the variance of parameters $\bvec{c}$ and $\bvec{W}$ except $\bvec{b}$ because $\bvec{b}$ is affected by many features of input patterns. Then inner product of variance of them should be monitor. 

\begin{figure}[tbp]
\begin{center}
\subfigure[Neuron Generation]{\includegraphics[scale=0.5]{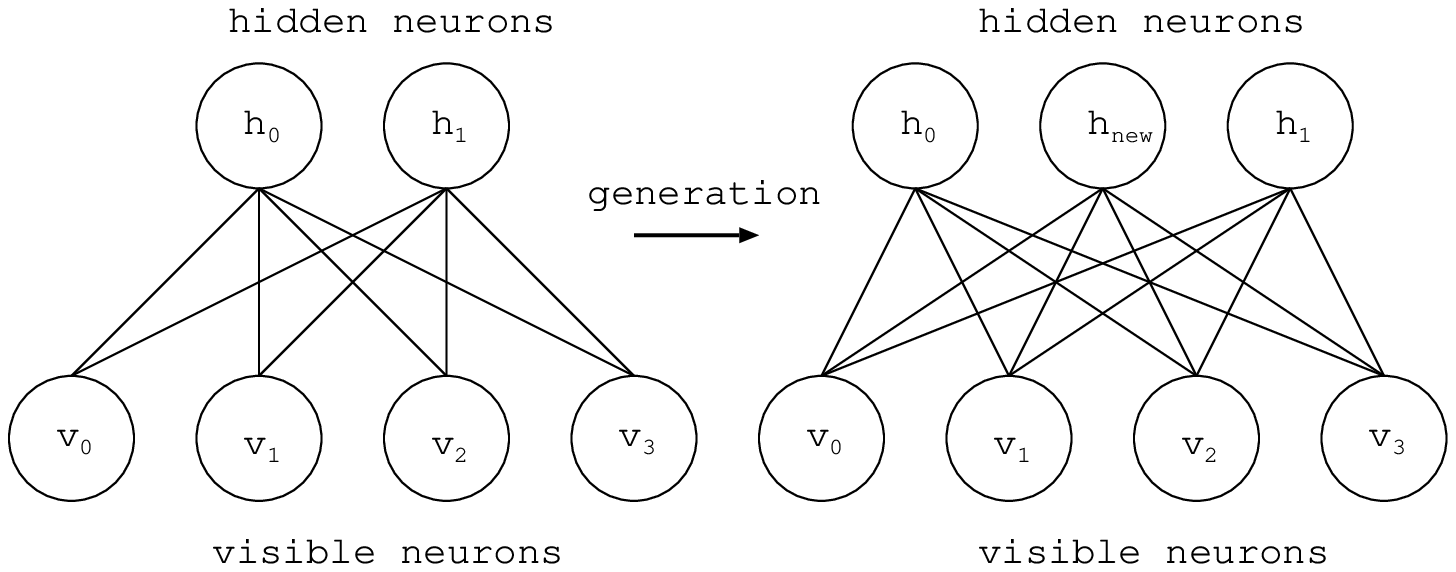}\label{fig:neuron_generation}}
\subfigure[Neuron Annihilation]{\includegraphics[scale=0.5]{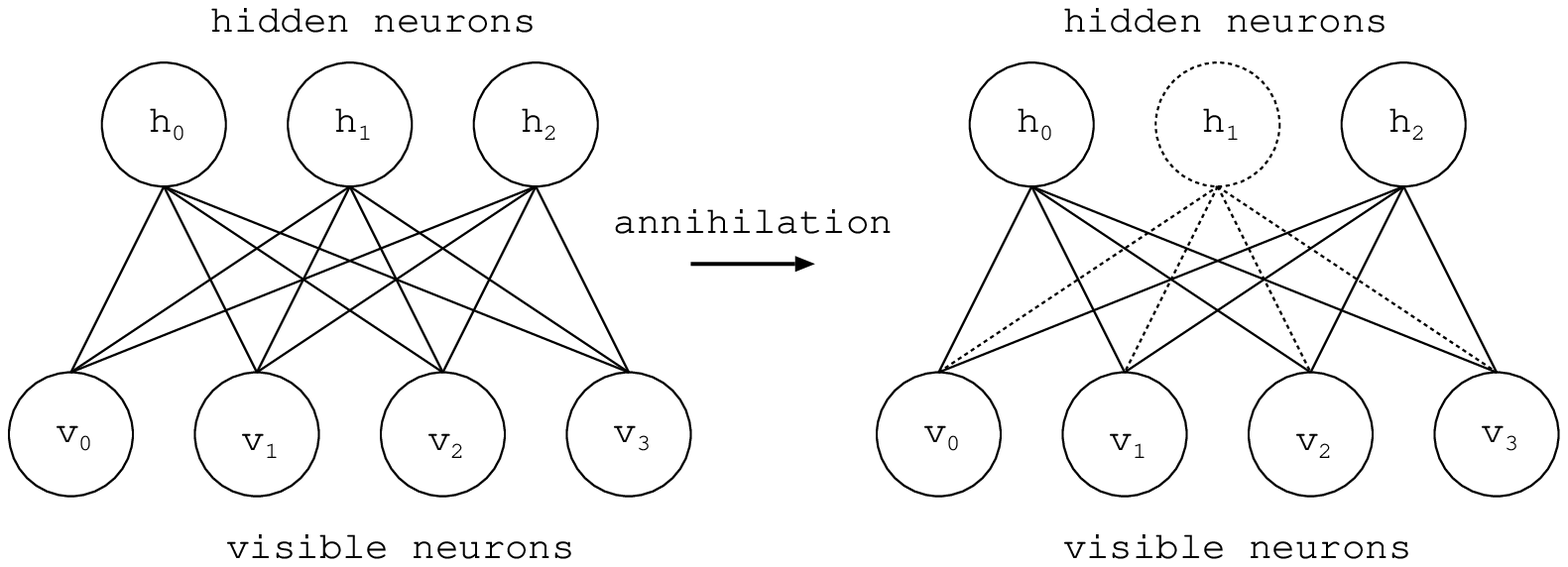}\label{fig:neuron_annihilation}}
\vspace{-3mm}
\caption{Adaptive RBM}
\label{fig:adaptive_rbm}
\vspace{-3mm}
\end{center}
\end{figure}

For the reason, we define the condition of neuron generation in the proposed adaptive RBM as in Eq.(\ref{eq:neuron_generation}) without the gradients for $\bvec{b}$ .
\begin{equation}
\vspace{-2mm}
(\alpha_{c} \cdot dc_j) \cdot (\alpha_{W} \cdot dW_{ij} )> \theta_{G},
\label{eq:neuron_generation}
\end{equation}
where $dc_j$ and $dW_{ij}$ are the gradient vectors of the hidden neuron $j$ and the weight vector $i, j$, respectively. $\alpha_{c}$ and $\alpha_{W}$ are the constant values for the adjustment of the range of each parameter. $\theta_{G}$ is an appropriate threshold value. The more $\theta_G$ is smaller, the more neuron generation is likely to be applied. A new hidden neuron is generated and is inserted into the neighborhood of the parent neuron as shown in Fig.~\ref{fig:neuron_generation}. The initial structure of RBM should be set arbitrary neurons according to the data set before training.

If some redundant neurons are generated, these neurons may be inactivated hidden neurons that does not contribute the classification capability. The proposed adaptive learning method of RBM determines the hidden neuron that satisfies with Eq.(\ref{eq:neuron_annihilation}), and then it annihilates the corresponding neuron as shown in Fig.~\ref{fig:neuron_annihilation}.
\begin{equation}
\vspace{-2mm}
\frac{1}{N}\sum_{n=1}^{N} p(h_j = 1 | \bvec{v}_n) < \theta_{A},
\label{eq:neuron_annihilation}
\end{equation}
where $\bvec{v}_n$ and $N$ are same as Eq.(\ref{eq:log_likelihood}). $p(h_j = 1 | \bvec{v}_n)$ means a conditional probability of $h_j \in \{ 0, 1 \}$ under a given $\bvec{v}_n$. $\theta_{A}$ is an appropriate threshold value. The more $\theta_A$ is higher, the more neuron annihilation is likely to be applied. Fig.~\ref{fig:neuron_annihilation} shows the structure of neuron annihilation. The dot circle is the redundant neuron removed by the annihilation algorithm.

\section{Experimental Results}
\label{sec:EXE}
This section describes experimental results to show the effectiveness of our proposed adaptive learning method of RBM.

\subsection{Data Set}
The 2 kinds of benchmark data set, ``MNIST \cite{mnist}'' and ``CIFAR-10 \cite{cifar10}'' were used in this experiments. MNIST is popular data set of handwritten digits and has 60,000 cases for training set and 10,000 cases for test set. They are categorized into 10 classes. Each case consists of $28 \times 28$ pixels. On the other hand, CIFAR-10 is 60,0000 color images data set included in 50,000 training cases and 10,000 test cases. They are categorized into 10 classes, and each case consists of $32 \times 32$ pixels. The original image data in CIFAR-10 only was preprocessed by ZCA whitening as reported in \cite{Coates11}.
 
In the experiments, Pylearn2 \cite{Goodfellow13}, which is one of machine learning tools with libraries for Deep Learning, was used for the implementation of RBM. The following parameters were used for RBM: the training algorithms is Stochastic Gradient Descent (SGD), the batch size is 100, and the learning rate is 0.1.

\subsection{Experimental Results}
\begin{figure*}[tbp]
\begin{center}
\subfigure[Energy Function]{\includegraphics[scale=0.56]{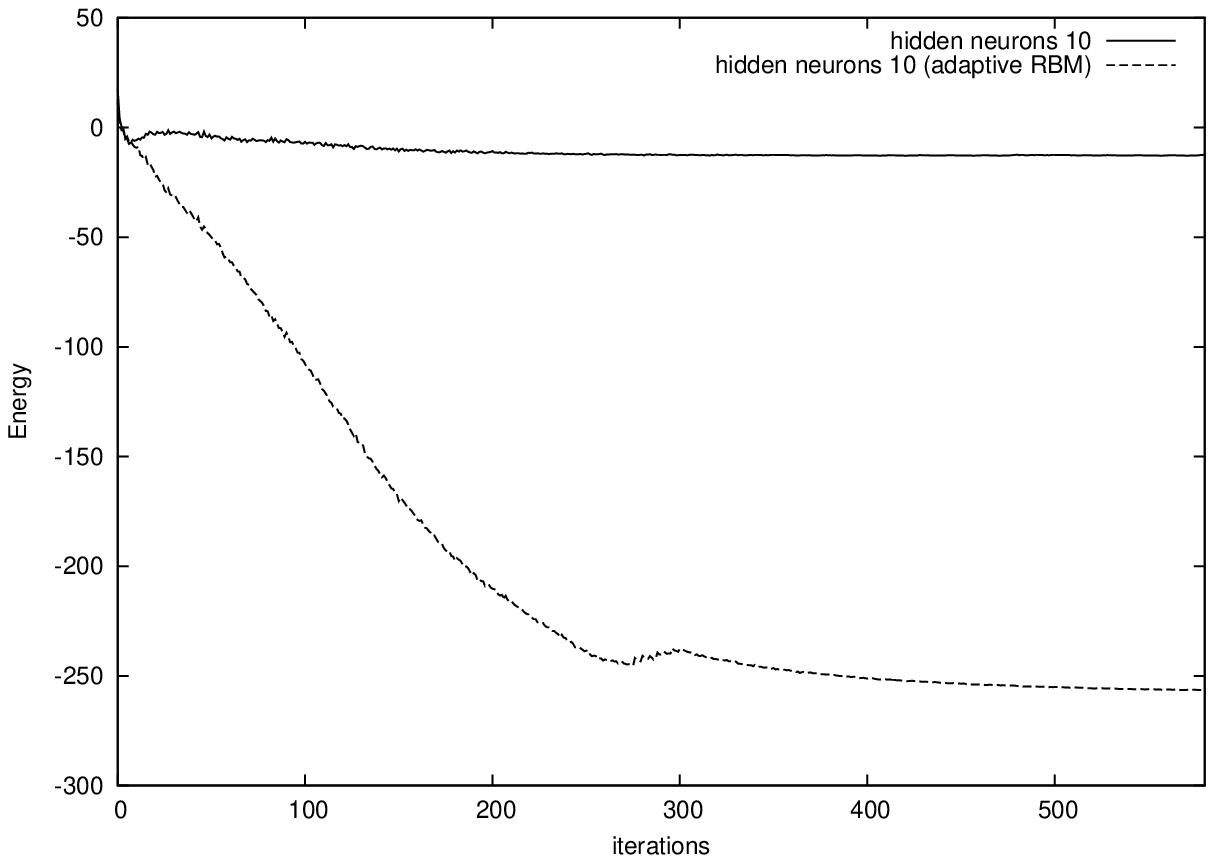}\label{fig:result-rbm-mnist-energy-10}}
\subfigure[Grad for each parameter (traditional RBM)]{\includegraphics[scale=0.56]{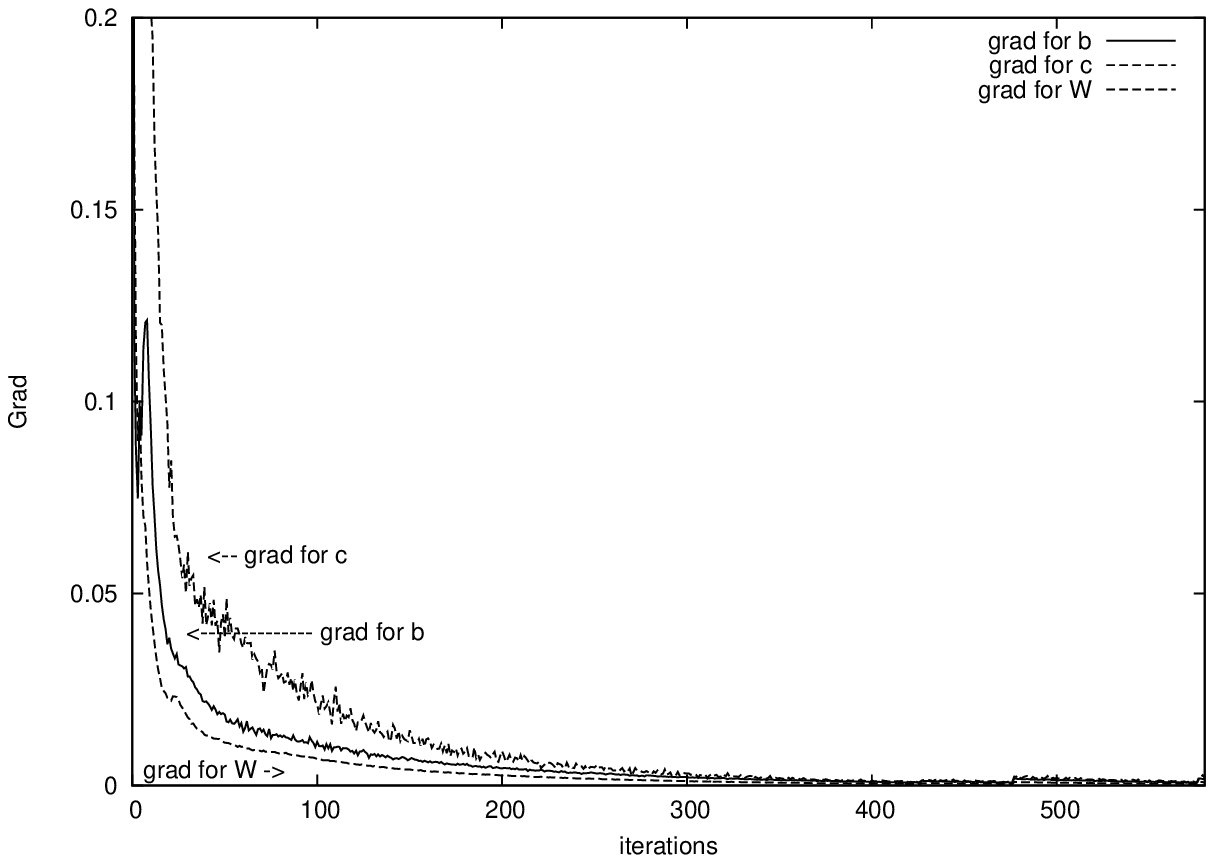}\label{fig:result-rbm-mnist-grad}}
\subfigure[Grad for each parameter (adaptive RBM)]{\includegraphics[scale=0.56]{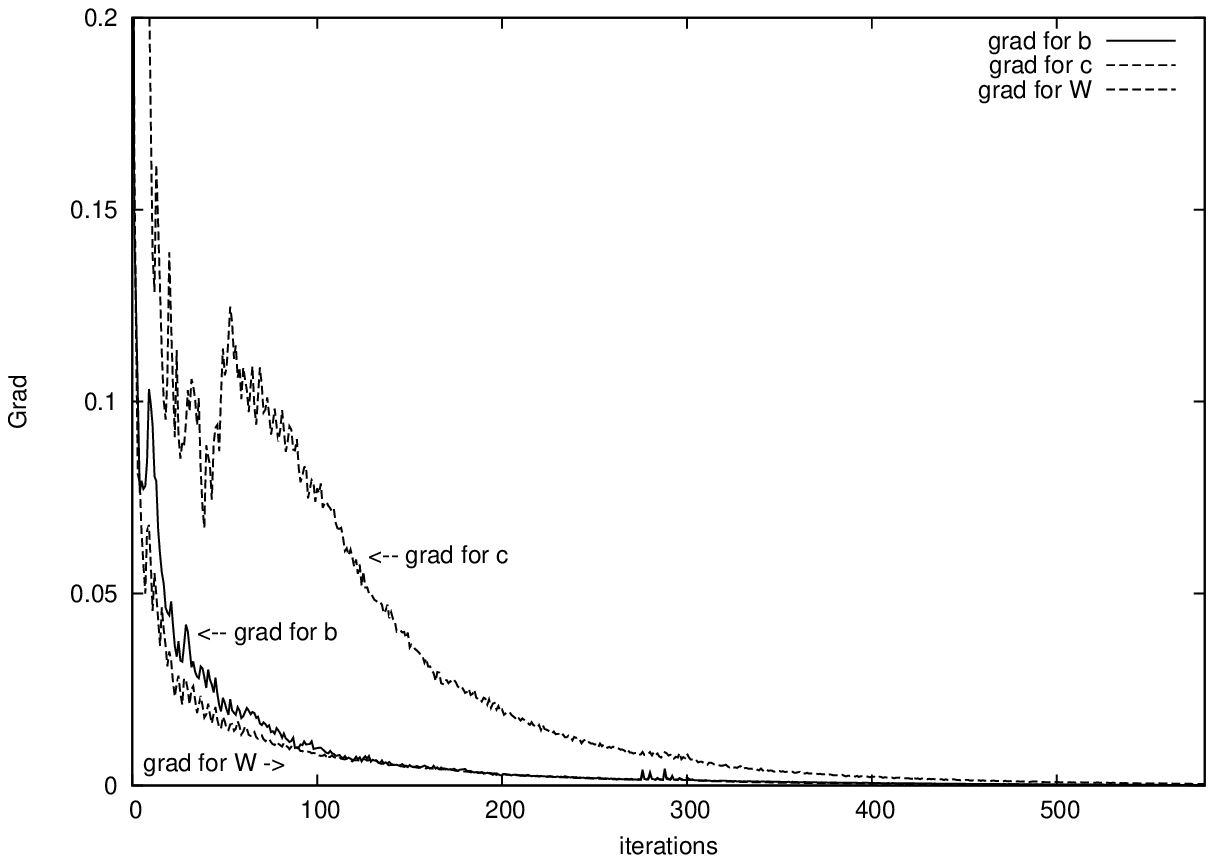}\label{fig:result-Arbm-mnist-grad}}
\subfigure[No. of hidden neurons]{\includegraphics[scale=0.56]{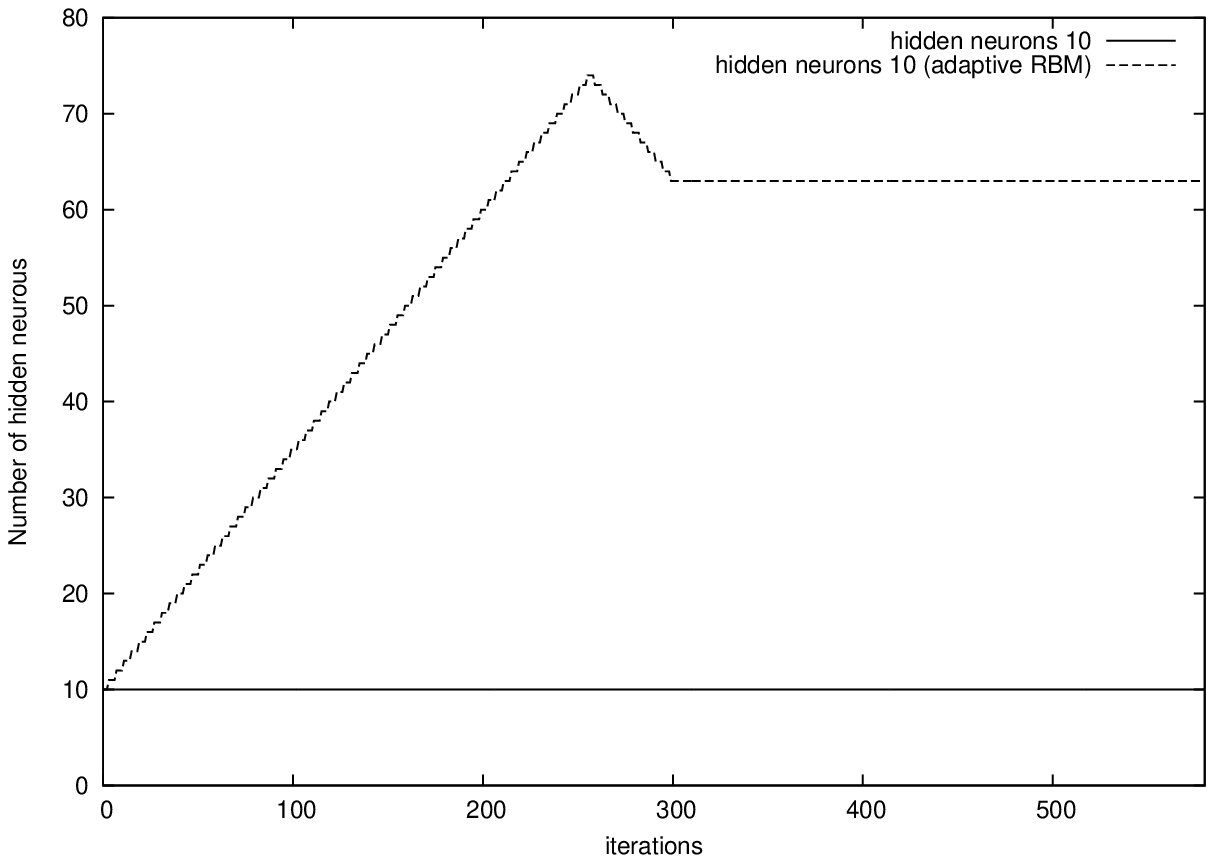}\label{fig:result-rbm-mnist-hnum-10}}
\end{center}
\vspace{-3mm}
\caption{MNIST (initial hidden neurons = 10)}
\label{fig:result_MNIST1}
\vspace{-3mm}
\end{figure*}

Fig.~\ref{fig:result_MNIST1} shows the experimental result for MNIST. Fig.~\ref{fig:result-rbm-mnist-energy-10} - ~\ref{fig:result-rbm-mnist-hnum-10} show the energy curve, the gradients for each parameter, and the number of hidden neurons. In this simulation, we set parameters as the following values: the initial number of hidden neurons is 10, $\theta_G = 0.005$, $\theta_A = 0.3$. The neuron generation algorithm starts to be applied from the 10th iteration because the learning situation is fluctuated at the beginning of learning due to the selection of initial parameters. About 60 additional hidden neurons were generated until about 250 iterations as shown in Fig.~\ref{fig:result-rbm-mnist-hnum-10}. A new generated neuron inherited the weight value from the parent hidden neuron, and it was inserted into the neighborhood of the parent. The proposed RBM became smaller energy than the traditional RBM without adaptation of network structure as shown in Fig.~\ref{fig:result-rbm-mnist-energy-10}.

After about 250 iterations, the neuron annihilation algorithm was worked as shown in Fig.~\ref{fig:result-rbm-mnist-hnum-10}, then the redundant neurons were removed from 71 to 62 gradually. The energy curve and the gradients for each parameter were not affected by the annihilation process.

\begin{figure*}[tbp]
\begin{center}
\subfigure[Energy Function]{\includegraphics[scale=0.56]{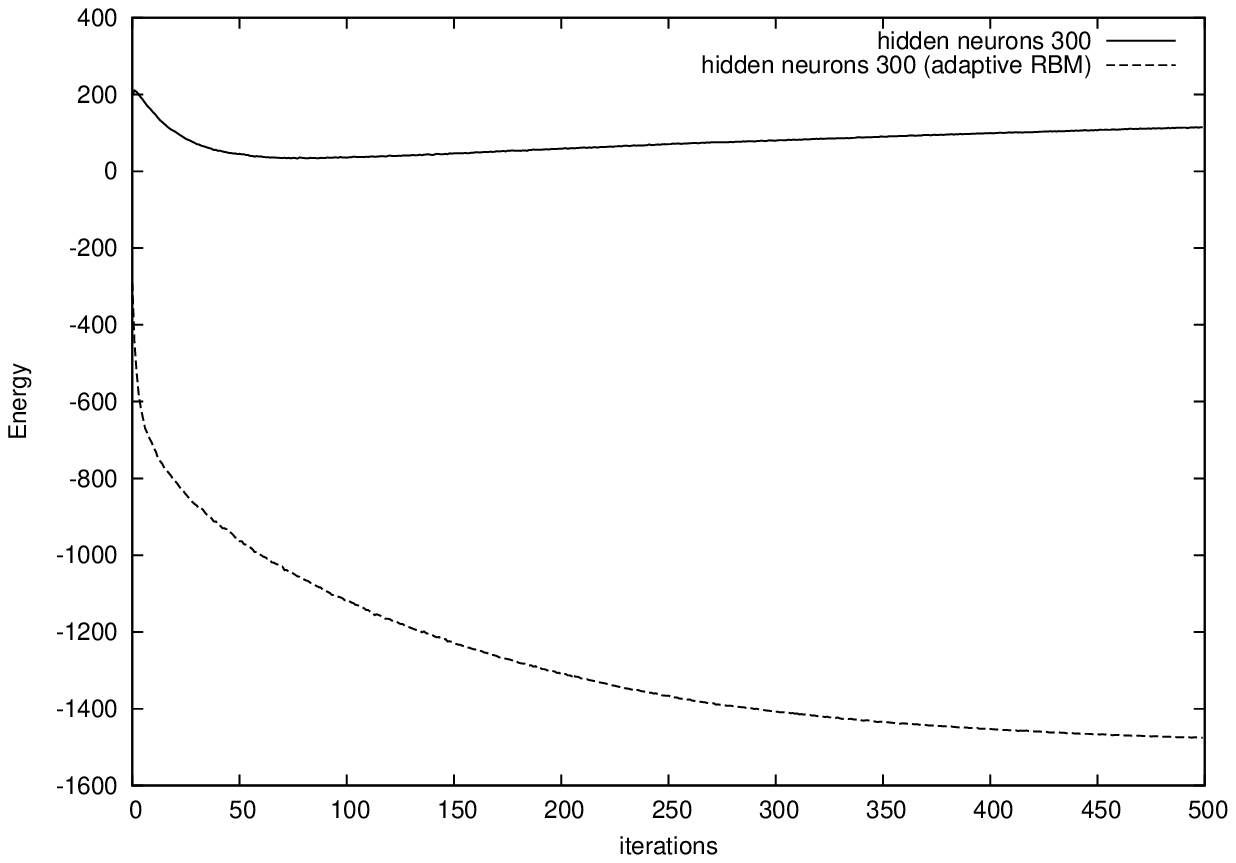}\label{fig:result-rbm-cifar10-energy}}
\subfigure[Grad for each parameter (traditional RBM)]{\includegraphics[scale=0.56]{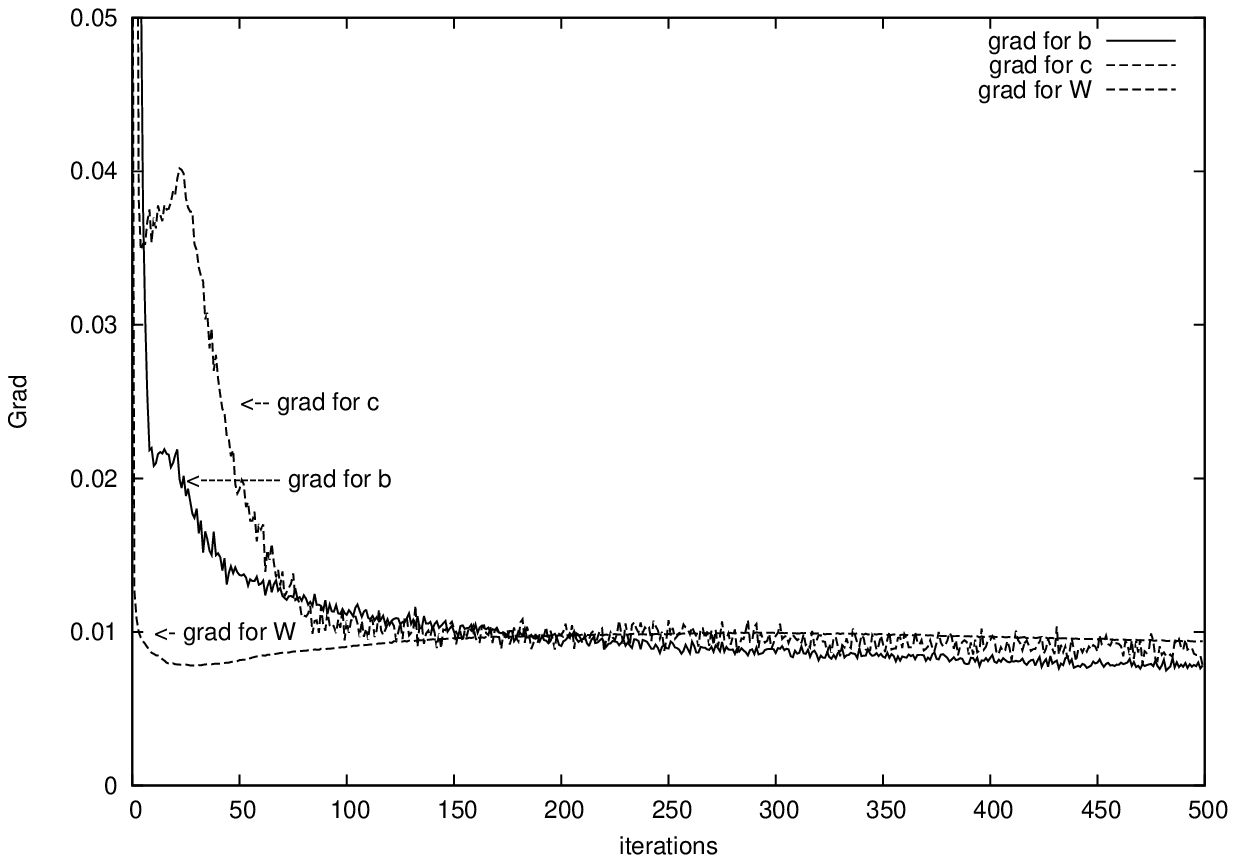}\label{fig:result-rbm-cifar10-grad}}
\subfigure[Grad for each parameter (adaptive RBM)]{\includegraphics[scale=0.56]{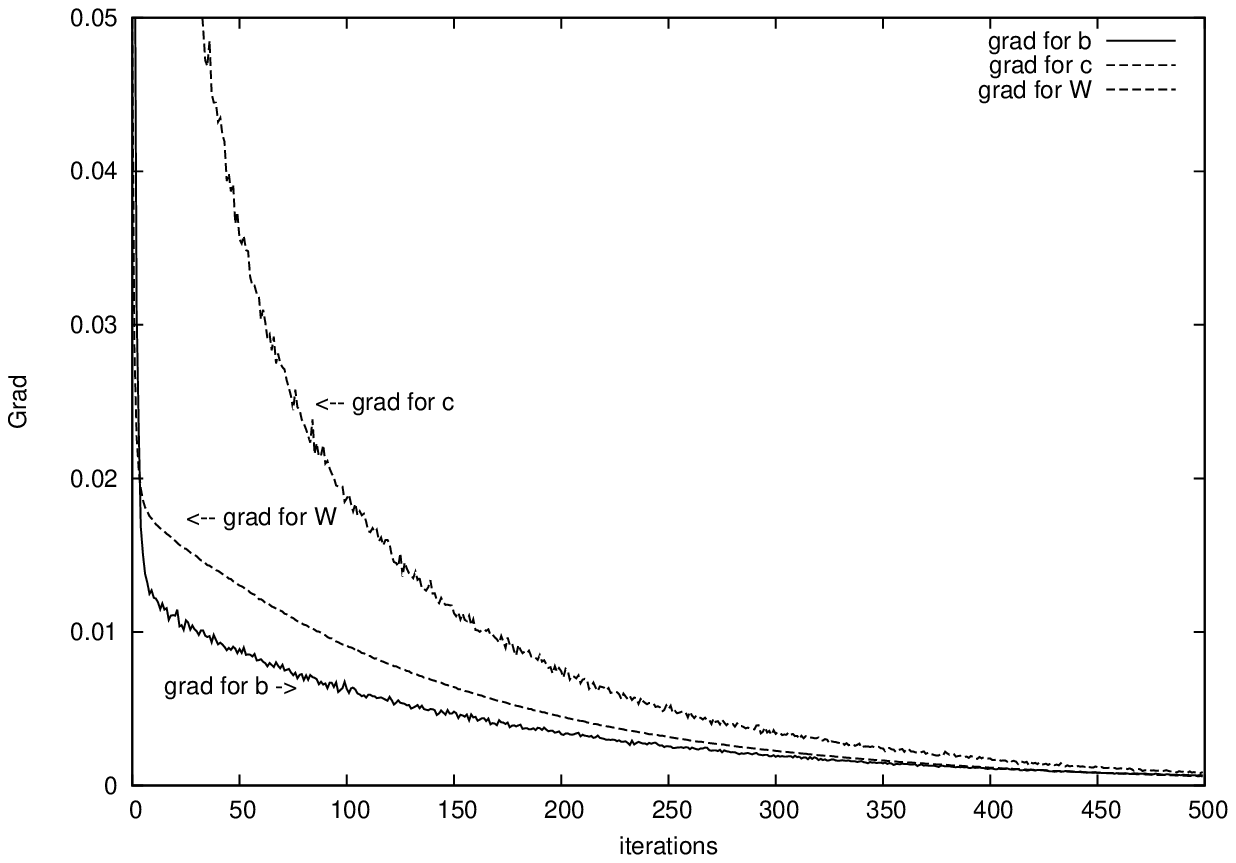}\label{fig:result-Arbm-cifar10-grad}}
\subfigure[No. of hidden neurons]{\includegraphics[scale=0.56]{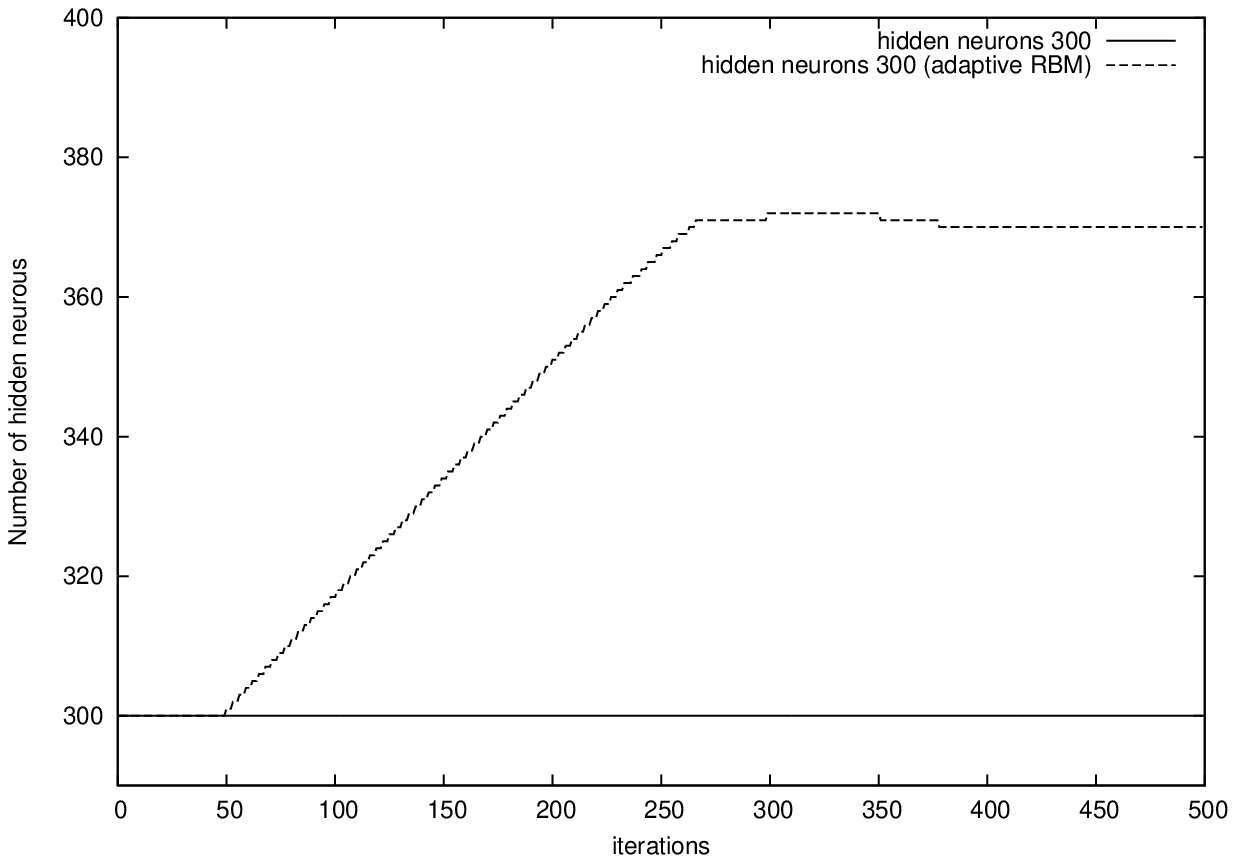}\label{fig:result-rbm-cifar10-hnum}}
\end{center}
\vspace{-3mm}
\caption{CIFAR-10 (initial hidden neurons = 300)}
\label{fig:result_CIFAR10}
\vspace{-3mm}
\end{figure*}

Fig.~\ref{fig:result_CIFAR10} shows the experimental results for CIFAR-10. We set parameters as the following values for CIFAR-10: the initial number of hidden neurons is 300, $\theta_G = 0.015$, $\theta_A = 0.01$. The experimental results for CIFAR-10 showed the same characteristics as the result of MNIST. The fluctuated gradients for each parameter in the traditional RBM was higher in comparison to the result of MNIST as shown in Fig.~\ref{fig:result-rbm-mnist-grad} and Fig.~\ref{fig:result-rbm-cifar10-grad} because CIFAR-10 has more ambiguous features of input patterns. On the other hand, the number of hidden neurons of our proposed RBM model was about 370 after the neuron generation. After the generation process, and the annihilation algorithm was implemented as shown in Fig.~\ref{fig:result-rbm-cifar10-hnum}. As a result, energy curve and gradients for each parameter at last iteration were converged with smaller value than the traditional RBM as shown in Fig.~\ref{fig:result-rbm-cifar10-energy} - ~\ref{fig:result-Arbm-cifar10-grad}.

In order to evaluate the classification capability with trained RBM, Hinton introduced 2 or more kinds of methods \cite{Hinton02}. As the evaluation method for the experiment, the output layer for classification is added into the trained hidden layer of RBM, then the network between them is fine-tuned by using BP learning. Table~\ref{tab:result-correct-ratio-MNIST} shows the classification accuracy for 10 kinds of images in MNIST. The classification accuracy of the proposed RBM was higher than the traditional RBM not only for training set but also for test set. Table~\ref{tab:result-correct-ratio-CIFAR10} shows the classification accuracy for 10 kinds of images in CIFAR-10. The traditional RBM didn't classify correctly similar classes such as 'dog' and 'cat' in CIFAR-10. On the other hand, such misclassification rate was decreased in the proposed adaptive RBM because some neurons which can represent ambiguous pattern were generated. As a result, the proposed RBM showed higher classification capability compared with the result of another RBM model reported in \cite{Dieleman12} (see \cite{Dieleman12}, for details for another RBM). 

\begin{table}[tbp]
\caption{Classification Accuracy on the MNIST}
\vspace{-5mm}
\label{tab:result-correct-ratio-MNIST}
\begin{center}
\begin{tabular}{r|r|r}
\hline \hline
                                       &  training set & test set\\ \hline\hline
traditional RBM (hidden neurons = 10) &  93.4\%  & 72.9\% \\ \hline
adaptive RBM (hidden neurons = 77)    &  100.0\% & {\bf 83.3}\%  \\ \hline
\hline 
\end{tabular}
\end{center}
\end{table}
\begin{table}[tbp]
\caption{Classification Accuracy on the CIFAR-10}
\vspace{-5mm}
\label{tab:result-correct-ratio-CIFAR10}
\begin{center}
\begin{tabular}{r|r|r}
\hline \hline
                                                   &  training set & test set  \\ \hline\hline
traditional RBM (hidden neurons = 300)             &  99.9\% & 70.1\% \\ \hline
adaptive RBM (hidden neurons = 370)                &  99.9\% & {\bf 81.2}\%  \\ \hline
sparce RBM (hidden neurons = 200)\cite{Dieleman12} &  - & 63.0\%  \\ \hline
\hline 
\end{tabular}
\vspace{-5mm}
\end{center}
\end{table}

\section{Conclusion}
\label{sec:Conclusion}
The problem related to the Deep Learning is known to be the setting for many parameters. Especially, RBM has 3 kinds of parameters for visible and hidden neurons in addition to the weights between them. In this paper, we introduced how the learning of RBM is converged under the Lipschitz continuous condition, and monitored the variance of 3 kinds of parameters during learning. Based on such observations, this paper proposed the adaptive learning method of RBM that can discover an optimal number of hidden neurons according to the training situation by applying the neuron generation and annihilation algorithm. Especially, some parameters were used for the neuron generation condition. In the simulation, we verified the effectiveness of our proposed method with 2 kinds of benchmark data set. As a result, our proposed RBM showed the higher classification capability with stable energy in comparison to the traditional RBM. In order to improve the classification capability further, the deep architecture that can represent more multiple features of input patterns with hierarchical level such as Deep Belief Network \cite{Hinton06} is required. In future, we will develop the RBM learning method with such a hierarchical structure where an optimal number of hidden layers is automatically defined.

\section*{Acknowledgment}
This work was supported by JSPS KAKENHI Grant Number 25330366.

\end{document}